# Validation & Exploration of Multimodal Deep-Learning Camera-Lidar Calibration models

Venkat Karramreddy, Liam Mitchell
*Electrical and Computer Engineering, Michigan State University, East Lansing, MI, USA*
[karramre@msu.edu](karramre@msu.edu), mitch991@msu.edu

*Abstract—* **This article presents an innovative study in exploring, evaluating, and implementing deep learning architectures for the calibration of multi-modal sensor systems. The focus behind this is to leverage the use of sensor fusion to achieve dynamic, real-time alignment between 3D LiDAR and 2D Camera sensors. static calibration methods are tedious and time-consuming, which is why we propose utilizing Conventional Neural Networks (CNN) coupled with geometrically informed learning to solve this issue. We leverage the foundational principles of Extrinsic LiDAR-Camera Calibration tools such as RegNet, CalibNet, and LCCNet by exploring open-source models that are available online and comparing our results with their corresponding research papers. Requirements for extracting these visual and measurable outputs involved tweaking source code, fine-tuning, training, validation, and testing for each of these frameworks for equal comparisons. This approach aims to investigate which of these advanced networks produces the most accurate and consistent predictions. Through a series of experiments, we reveal some of their shortcomings and areas for potential improvements along the way. We find that LCCNet yields the best results out of all the models that we validated.**

## I. INTRODUCTION

In the realms of autonomous vehicles and robotics, perfecting the way computer vision applications operate can play a crucial role in mitigating accidents from occurring. While high-quality sensors can improve input accuracy, they may not necessarily enhance a computer's perception or interpretation of events. Beyond perfecting artificial intelligence training protocols, researchers have devised methods for correcting many of the elements that influence the quality of feed-forward input.

Achieving scientifically accurate and measurable results in the context of a modality is purely dependent on the methodology employed. Theoretically, modalities exist because they are human-made constructs that have been proven effective in interpreting and engaging with reality. Renowned 18th-century German philosopher Immanuel Kant famously stated, "We see the world and things not as they are, but as we are." Although insightful, many of nature's complex problems are ones we've presented to ourselves. However, recent advancements in deep learning, neural networks, and artificial intelligence have transformed our approach to these complexities, offering new perspectives and solutions. LiDAR and Cameras, essential CV sensors used for developing autonomous vehicles, robotics, and augmented reality, usually require manual calibration with fixed markers (environmental objects or checkerboard-pattern targets). These technologies rely heavily on knowing these measured calibration values so that systems can merge spatial and visual data accurately, a process critical for object detection, navigation, and environment mapping. Despite the critical importance of accurate multi-modal sensor fusion, achieving high levels of precision in the calibration process remains a challenging endeavor, for which the nature of dynamics can be unpredictable in real-world environments.

Significant research in 3D reconstruction calibration techniques has emerged through deep learning. Introducing a variety of smart, efficient, and quantifiable models that promote a streamlined, hands-free approach to calibration. As we delve further into this article's discussions, it's important to note that our specific expertise is not in artificial intelligence, but rather an extensive background in automotive R&D, particularly in sensor data acquisition, and ingestion processing. We assume our readers have the background knowledge necessary for understanding the rhetoric of the content we discuss. For those seeking a foundational or more in-depth understanding of camera calibration via deep learning, we highly recommend starting with: "Deep Learning for Camera Calibration and Beyond: A Survey" [16]. This paper, authored by specialists, offers a comprehensive review of recent advancements in deep learning calibration techniques.

## II. PROBLEM DESCRIPTION

Much of this information is dependent upon the fact that the designed neural networks have enough initial input parameters, which most of the time involve a large set of algorithms and functions that it calculates for making connections and learning.

As vehicles operate, vibrations may cause shifts in the position or orientation of sensors, necessitating constant adjustments in their calibration. Traditional manual calibration requires sensors to be recalibrated at regular intervals to ensure consistent and robust performance. In contrast, online calibration automates this process, continuously adjusting for any changes in sensor alignment.

By using geometry-based feature extraction techniques, we can establish a geometric correlation between 3D LiDAR point cloud data and 2D camera images, facilitating a qualitative representation of alignment accuracy through matching congruence of projected point clouds with respective features, edges, or objects in the camera images. Provided the datasets, real or simulated, originate from the same ego-vehicle with time-synchronized sensors, we can leverage an extensive collection of odometry data to accurately train a model's perception.

RegNet, CalibNet, LCCNet, and Calib-Anything are some of the methods that use deep learning methods to tackle extrinsic calibration and have been shown to have accurate results. However, when we independently evaluated these models under identical experimental conditions using the KITTI dataset, we found that there is a notable difference between our findings and the results published in those papers. Our findings show that the mean translational and rotational errors are 3 to 10 times worse than the original results.

*Extrinsic Calibration Background information:*

When a camera's image is analyzed, it can be considered to have several different coordinate systems. For instance, any 3D object that is in the world coordinate system is converted to the camera coordinate system and from the camera coordinate system to the image coordinate system. And finally, from the image coordinate system to the pixel coordinate system.

For the conversion from world coordinate system to the camera coordinate system, only the values in the x-axis and y-axis direction will change when the system is rotated by θ along the z-axis. Similarly, when the coordinate system is simultaneously rotated by ω along the x-axis, rotated by ψ along the y-axis, the rotation matrix (denoted as R) is:

$$R = \begin{bmatrix} \cos\theta & -\sin\theta & 0 \\ \sin\theta & \cos\theta & 0 \\ 0 & 0 & 1 \end{bmatrix} \times \begin{bmatrix} 1 & 0 & 0 \\ 0 & \cos\omega & -\sin\omega \\ 0 & \sin\omega & \cos\omega \end{bmatrix} \times \begin{bmatrix} \cos\psi & -\sin\psi & 0 \\ \sin\psi & \cos\psi & 0 \\ 0 & 0 & 1 \end{bmatrix}$$

$$= \begin{bmatrix} R_{11} & R_{12} & R_{13} \\ R_{21} & R_{22} & R_{23} \\ R_{31} & R_{32} & R_{33} \end{bmatrix}$$

When the coordinate system is moved along the x, y, and z axis then the translation matrix is:

$$T = \begin{bmatrix} T_x \\ T_y \\ T_z \end{bmatrix}$$

Therefore, we can obtain the homogeneous coordinates for the transformation from the world coordinate system ($x_w$, $y_w$, $z_w$) to the camera coordinate system ($x_c$, $y_c$, $z_c$) as follows:

$$\begin{bmatrix} x_c \\ y_c \\ z_c \\ 1 \end{bmatrix} = \begin{bmatrix} R_{11} & R_{12} & R_{13} & t_x \\ R_{21} & R_{22} & R_{23} & t_y \\ R_{31} & R_{32} & R_{33} & t_z \\ 0 & 0 & 0 & 1 \end{bmatrix} = \begin{bmatrix} R & T \\ 0 & 1 \end{bmatrix} \begin{bmatrix} x_w \\ y_w \\ z_w \\ 1 \end{bmatrix}$$

where R refers to the rotation matrix and T refers to the translation matrix, which is the transformation between the world coordinate system and the camera coordinate system.

Secondly, the relationship between the image coordinate system and the pixel coordinate system is:

$$\begin{bmatrix} u \\ v \\ 1 \end{bmatrix} = \begin{bmatrix} \frac{1}{dx} & 0 & u_0 \\ 0 & \frac{1}{dy} & v_0 \\ 0 & 0 & 1 \end{bmatrix} \begin{bmatrix} x \\ y \\ 1 \end{bmatrix}$$

Finally, the transformation relationship between the world coordinate system and the pixel coordinate system is:

$$Z_c \begin{bmatrix} u \\ v \\ 1 \end{bmatrix} = \begin{bmatrix} f_x & 0 & u_0 & 0 \\ 0 & f_y & v_0 & 0 \\ 0 & 0 & 1 & 0 \end{bmatrix} \begin{bmatrix} R & T \\ 0 & 1 \end{bmatrix} \begin{bmatrix} x_w \\ y_w \\ z_w \\ 1 \end{bmatrix}$$

$$K_1 = \begin{bmatrix} f_x & 0 & u_0 & 0 \\ 0 & f_y & v_0 & 0 \\ 0 & 0 & 1 & 0 \end{bmatrix}$$

$$K_2 = \begin{bmatrix} R & T \\ 0 & 1 \end{bmatrix}$$

where K2 is the external parameter of the camera, which is also a parameter to be solved, closely related to the relative position of the camera and the lidar, and K1 is the internal parameter of the camera, which is only related to the interior of the camera.

### III. THEORETIC

*III-A. RegNet:*

RegNet [6] works on the principle of regressing 6 DOF for extrinsic calibration by leveraging deep neural networks for feature extraction and feature matching [1] [2] [3]. The problem of extrinsic calibration is reformulated as determining the decalibration $\varphi_{decalib}$ given the initial calibration $H_{init}$ and a ground truth calibration $H_{gt}$.

Given initial extrinsic $H_{init}$ and camera intrinsic $K$, the depth image is generated by projecting 3D LiDAR point cloud from the LiDAR scan onto a virtual image plane with a 2D pixel coordinate $(u, v)$ using the formula (1). Where $z_c$ is the inverse depth of the projected point and $x$ is the $(x, y, z)$ of the lidar point. The homogeneous calibration matrix is composed of a 3x3 rotation matrix $R$ and a 3x1 translation vector $t$ (2).

$$z_c \begin{bmatrix} u \\ v \\ 1 \end{bmatrix} = K * H_{init} \, x \qquad (1)$$

$$\phi_{decalib} = \begin{bmatrix} R & t \\ 0 & 0 & 0 & 1 \end{bmatrix} \qquad (2)$$

*Architecture:*

The network is designed as an end-to-end CNN model which can solve the task of feature extraction, feature matching, and regression of the calibration parameters by using several Network-in-Network (NiN) blocks [4] as shown in Figure 1. The features are extracted by giving RGB images and LiDAR depth maps as network inputs in parallel streams. The architecture and weights of the NiN [15] for ImageNet [5] is used for the RGB part, while the depth stream is learned from scratch. Then the output feature maps are concatenated and convolved through a series of NiN blocks to match the features. Finally, a Euclidean loss function is used to optimize the extrinsic calibration parameters.

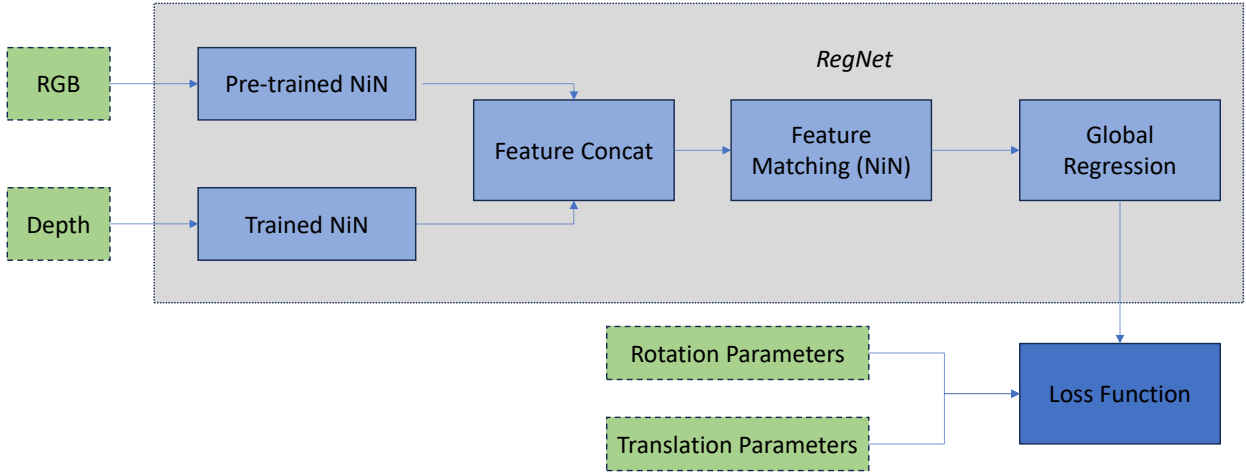

Figure 1: RegNet architecture

*Training:*

For the RegNet model we used, the only way for results to be extract was to use its inference module, which required training. During its training we noticed the file ingestion portion of the code required tweaking for it's nueral network to learn properly. You can see in Figure 2, proper loss was achieved.

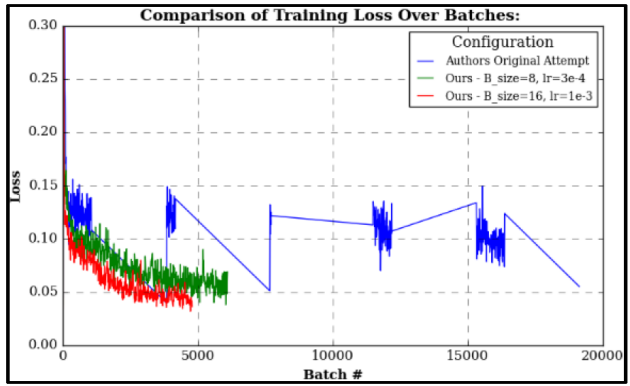

Figure 2: Training Loss fix for RegNet model we used.

Figure 3, shows the training results of the RegNet calibration after the initial decalibration and also the comparison with the Ground Truth data.

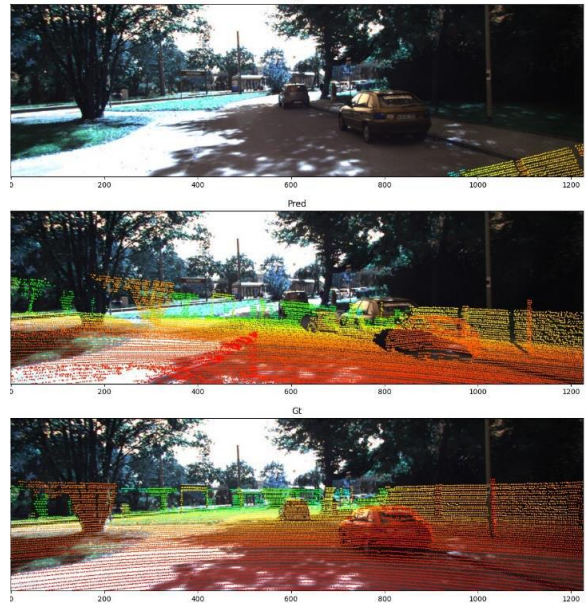

Figure 3: Visual Results from RegNet models test

*III-B. CalibNet:*

CalibNet's [7] architecture is based on 3D Spatial Transformers [8] and uses geometric and photometric consistency as a loss function to learn the external parameters.

*Architecture:*

The network takes the RGB image as one input and the LiDAR depth map, which is obtained by projecting the LiDAR points onto the 2D image plane, as another input (1). Since the initial calibration is inaccurate, the projected points on the depth image are inconsistent with the image. CalibNet uses 2 parallel branches to extract the features from RGB images and depth images. It uses a pre-trained ResNet-18 [9] network to extract features from RGB images. For the depth branch, it uses a similar architecture but with half the number of filters, as it needs to learn from scratch. The outputs of the two branches are then concatenated and then passed through a series of convolutional layers for global feature aggregation. The outputs from the global feature aggregation are a translational vector and a rotational velocity vector. Using the Rodrigues formula, the rotational vector is converted to a rotation matrix. This rotation matrix R combined with the translation vector t gives a 3D rigid body transformation T.

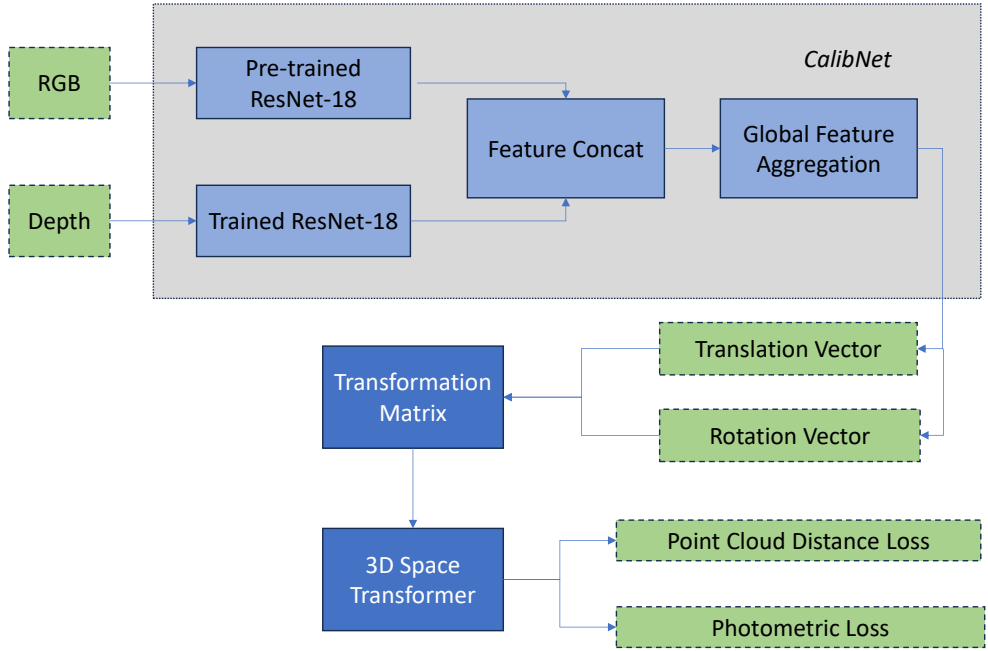

Figure 4: CalibNet architecture

$$T = \begin{bmatrix} R & t \\ 0 & 1 \end{bmatrix} \qquad (3)$$

CalibNet uses a 3D Spatial Transformer Layer to transform the input depth map into a sparse point cloud using the predicted transformation $T$ and camera intrinsics $(f_x, f_y, c_x, c_y)$ (4). Then the transformed point cloud is projected back into the image plane using the extrinsic calibration $T$ (5).

$$K^{-1}(x, y, Z) = \left( \left( \frac{x - c_x}{f_x} \right), \left( \frac{y - c_y}{f_y} \right), Z \right) \qquad (4)$$

$$\begin{pmatrix} x \\ y \end{pmatrix} = K \left( R \begin{pmatrix} X \\ Y \\ Z \end{pmatrix} + t \right) \qquad (5)$$

A photometric loss function is calculated by using the dense pixel-wise error between the depth map by the predicted $T$, and the ground truth depth map. Similarly, a point cloud distance loss is also calculated between the predicted depth map and the ground truth depth map. A weighted sum of the photometric loss and point cloud distance loss is used as the loss function. The newly transformed depth map can be used as an input to the model to refine the calibration. This process can be done multiple

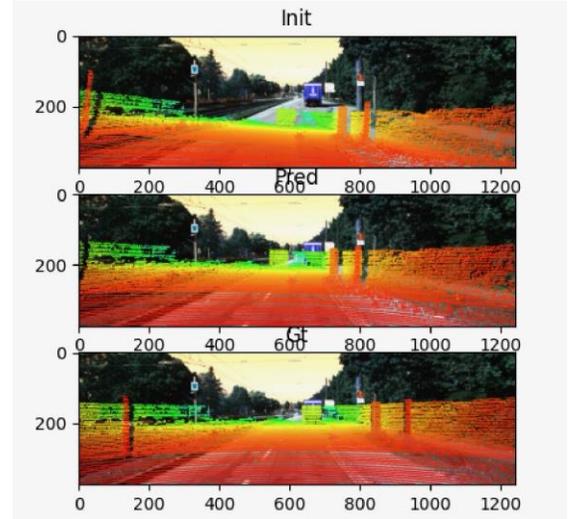

Figure 5: Results from CalibNet model

times to further refine the external calibrations.

The above Figure 5, shows the predictions of the point cloud when used with CalibNet model, in comparison with the initial decalibration and the Ground Truth point clouds.

### III-C. LCCNet:

LCCNet is an end-to-end learning network comprising of feature extraction layer, feature matching layer, and global feature aggregation network to calibrate the extrinsic parameters between LiDAR and Camera. It uses a smooth $L_1 - loss$ and point cloud distance loss as loss function, between the predicted and actual calibration.

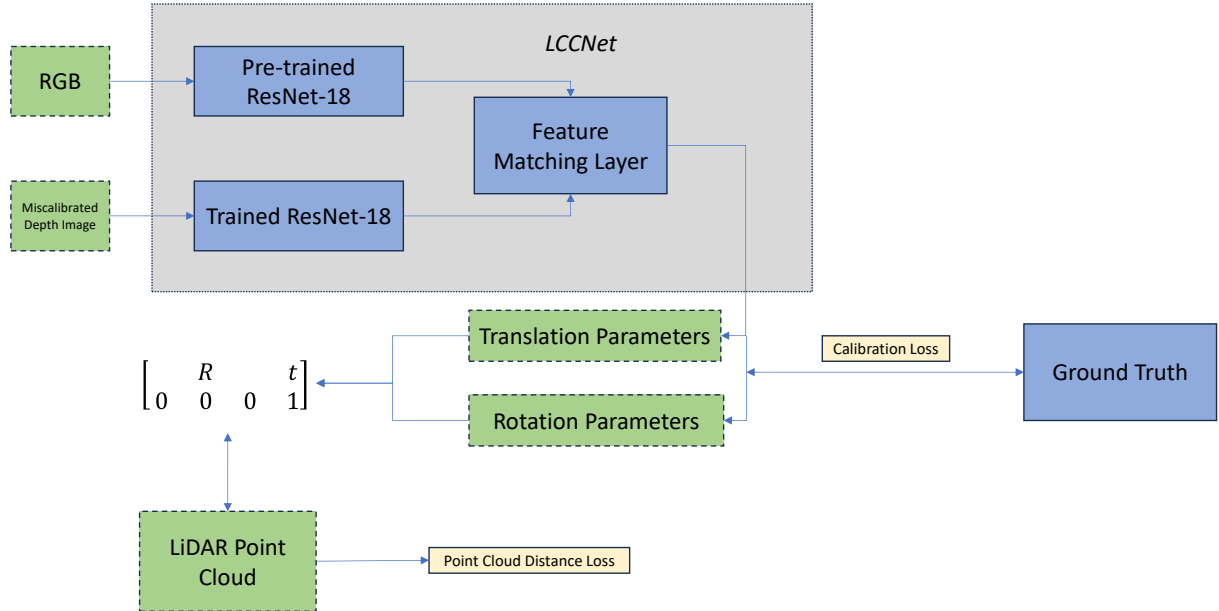

Figure 6: LCCNet architecture

*Architecture:*

Like CalibNet, LCCNet also converts LiDAR point cloud into depth map (1) and uses two parallel streams for feature extraction. A pre-trained ResNet-18 for RGB image and a similar branch for depth map. Unlike CalibNet, it uses a Leaky RELU as the activation function. After feature extraction, a cost function is used to calculate the associating cost of pixel in RGB feature maps $x_{rgb}$ with its corresponding depth feature maps $x_{lidar}$ [12] (6).

$$cv(p_1, p_2) = \frac{1}{N}\Big(c\big(x_{rgb}(p_1)\big)\Big)^T c\big(x_{lidar}(p_2)\big) \quad (6)$$

where $c(x)$ is the flattened vector of feature maps $x$, $N$ is the length of the column vector $c(x)$. For feature matching, the local cost volume is limited to 2 pixels. Then the network uses a weighted sum of regression loss $L_T$ and point cloud distance loss $L_P$ as a loss function (7) where regression loss is the smooth $L_1$ loss between the ground truth quaternions and prediction quaternions, and the point cloud distance loss is the $L_2$ normalization of the distance between the predicted point cloud and the ground truth point cloud.

$$L = \lambda_T L_T + \lambda_P L_P \quad (7)$$

For further refinement of the calibration parameters, the initial prediction $T_{pred}$ is given as $T_{init}$ to the network predict new transformation. This process can be repeated few times to further refine the values.

*III-D. Calib-Anything:*

Calib-Anything [14] uses Segment Anything Module (SAM) [13] and point cloud consistency to calculate the extrinsic calibration parameters. It optimizes for intensity, normal vector, and segmentation class of point cloud on masks.

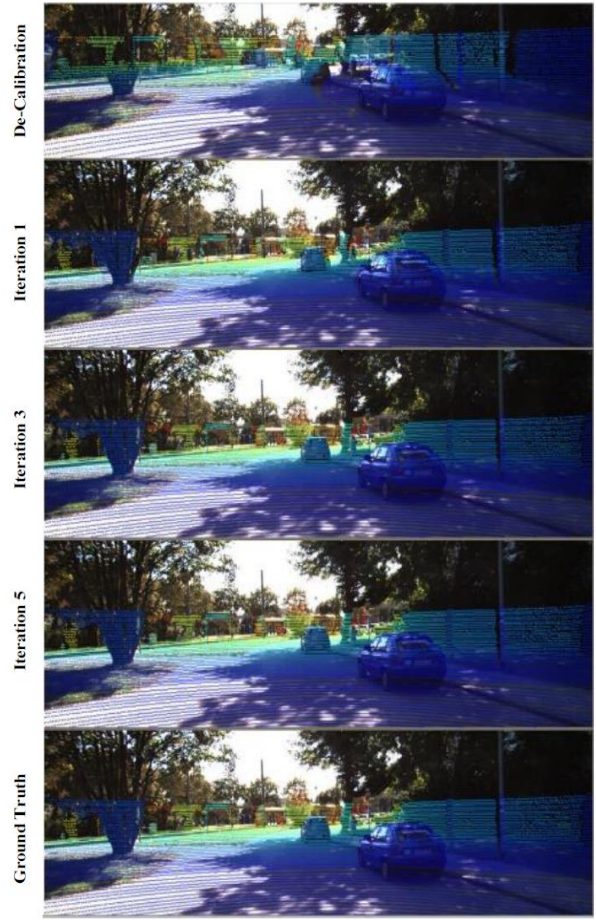

Figure 7. Results from LCCNet models test.

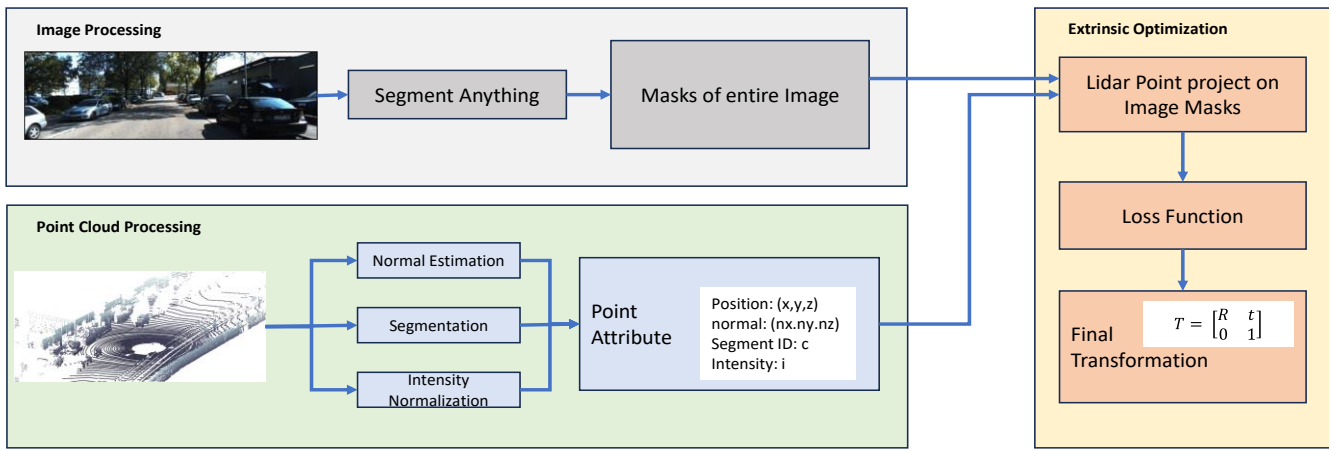

Figure 8: Calib-Anything architecture

*Architecture:*

Calib-Anything involves three main steps: image segmentation using SAM, point cloud preprocessing, and extrinsic optimization. SAM is first applied on the entire image yielding multiple masks; and the hyperparameters of SAM are adjusted to obtain more masks with less overlapping areas. Each mask is a binary matrix of the same size as the image, where each pixel denotes if it belongs to a segment or not.

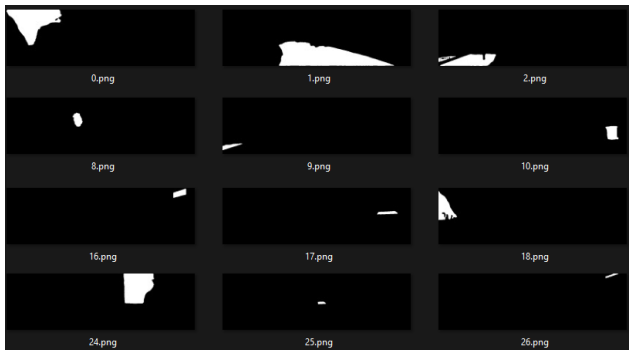

Figure 9: SAM Generating Segmentation Masks

Point cloud preprocessing step contains three parts - normal estimation which involves using Principal Component Analysis (PCA) on a covariance matrix generated from nearby neighbors of the query point. Intensity normalization to account for a range of point cloud intensities. And segmentation, to get clusters of individual objects like vehicles and trees. The final attribute of appoint in the point cloud is represented as:

$$p = \{x, y, z, n_x, n_y, n_z, r, c\} \quad (8)$$

which is the position, normal vector, reflectivity, and segmentation class of the point P.

Lastly for the extrinsic optimization, the LiDAR point cloud is projected on the image plane and a consistency score is calculated for each point falling on each mask. Reflectivity consistency is calculated by the standard deviation of the corresponding point values. Normal vector consistency is calculated by the pairwise dot products of the vectors and segmentation consistency is calculated by the weighted sum of all classes. The final consistency score is the weighted mean of the scores of all masks. A brute force method is applied on the rotation values, with a large step size, within certain limits of the initial calibration, to calculate the best consistency score. Then finer step sizes are used on all 6 extrinsic parameters to get more refined extrinsic calibration parameters.

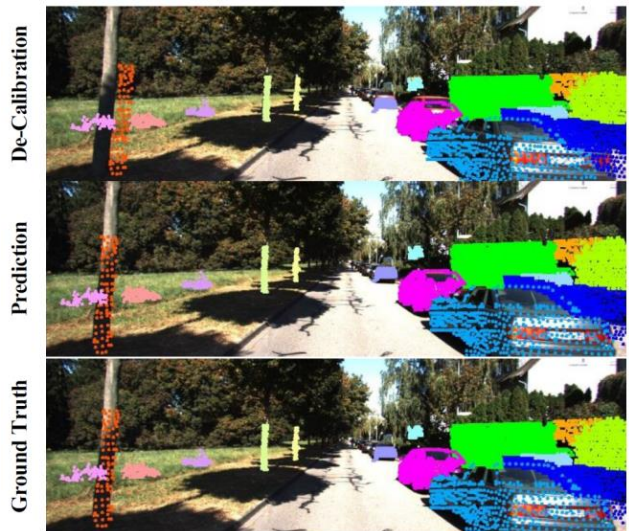

Figure 10: Results from Calib-Anything models test

Calib-Anything is an interesting model. From its source code, it appeared to condense the point cloud into voxels, thus reducing sparse outliers, which is what you see overlayed in Figure 10 above. With the condensed point cloud voxels it made it much easier for the model to match features from the image's segmented masks so it could make predictions for the Extrinsics. It is important to note that although it did not require training, it did take much longer than any of the other models.

IV. NUMERICAL SIMULATION RESULTS

*A. Datasets:*

TABLE 1. QUICK REFERENCE OF OPEN-DATASETS AVAILABLE, THEIR SIZE, AND IF THEY WERE SELECTED OR USED FOR TESTING

| Dataset | Version (Downloaded) | Compressed (GB) | Uncompressed (GB) | Selected |
|---|---|---|---|---|
| KITTI | Visual Odom / SLAM Eval 2012 | 167 | 177 | X |
| KITTI | 3D Object Detection / 2017 | 39 | 42 | |
| nuScenes | v1.0-Trainval & v1.0-Test | 347 | 497 | |
| nuScenes | V1.0-mini | 5 | 5 | |
| ONCE | Train / Val / Test (Splits) | 62 | 66 | |
| Waymo | v1.2.0 | 49 | 102 | |

*KITTI (Visual Odometry / SLAM Evaluation 2012):*

Curated by the Karlsruhe Institute of Technology, the KITTI dataset is one of the most commonly used datasets which autonomous driving researchers and communities prefer. KITTI is acclaimed for its comprehensive suite of sensor data and has set the benchmark for evaluating 3D Object algorithms since its initial dataset was released in 2012. Although other datasets are available, per Table 1, most of these models have been created to only work with KITTI, which is why we used it for our evaluations.

When comparing these models against each other, we chose the most extreme de-calibration parameters we possibly could. With a Rotation of ±25° and Translation of 1.5 meters. We were quickly able to see which open-source tools set themselves apart in both positive and very negative ways. Results for extracted Extrinsic values is shown below in Table 2.

*B. Results:*

TABLE 2. MEASURABLE RESULTS EXTRACTED FROM MODELS

| | Authors Initial Decalibration (R / T) | Authors Mean Translational Error (cm) | Authors Mean Rotational Error (degrees) | Our Initial Decalibration ( R / T ) | Our Mean Translational Error (cm) | Our Mean Rotational Error (degrees) |
|---|---|---|---|---|---|---|
| RegNet | 20° / 1.5m | 6 | 0.28 | 25° / 1.5m | 83 | 4.28 |
| CalibNet | 20° / 0.2m | 4.34 | 0.41 | 25° / 1.5m | 4.99 | 4.065 |
| LCCNet | 25° / 1.5m | 0.297 | 0.017 | 25° / 1.5m | 0.9588 | 0.0972 |
| Calib-Anything | 10° / 0.5m | 10.4 | 0.168 | 10° / 0.5m | 26 | 0.77 |

TABLE 3. VISUAL RESULTS EXTRACTED FROM MODELS

| | RegNet | CalibNet | LCCNet | Calib-Anything |
|---|---|---|---|---|
| Decalibration | | | | |
| Prediction | | | | |
| Ground Truth | | | | |

## V. CONCLUSION

The most difficult task, apart from finding open-source tools, was getting these models to compile. Most of them were research projects without the intention of being maintained and have since become deprecated. The others are questionable in how they produce their results. Many, even with pre-trained weights, would initially only generate Ground-Truth Projections instead of what might be perceived as a fully functioning software given the authors' published results. Despite these challenges, we were able to compile and validate a visually distinct list for 4 of the extrinsic calibration tools. Among the evaluated tools, we found that LCCNet had the most accurate and consistent performance, with Calib-Anything following closely. Future research should focus on developing more adaptive and scalable calibration frameworks that can handle diverse operational contexts and dataset variations. Although newer and more robust models exist, it is difficult to evaluate their results independently due to open-source code frequently being stolen and reused by other researchers, which is a very unfortunate reality. By continuing to refine these technologies, we can significantly enhance the reliability and safety of AVs and their multimodal sensor systems, pushing the boundaries of what is possible with artificial intelligence and deep learning models for real-world applications.